%% file: paper.tex
\documentclass[conference]{IEEEtran}
\usepackage{times}

\usepackage[numbers, sort]{natbib}
\usepackage{multicol}
\usepackage{multirow}
\usepackage{tabularx}
\usepackage[bookmarks=true]{hyperref}

\usepackage{amsmath}
\usepackage{xfrac}
\usepackage{amssymb}
\usepackage{amsfonts}
\usepackage{mathtools, nccmath}
\usepackage[dvipsnames,svgnames]{xcolor}

\usepackage{siunitx}
\usepackage[capitalize]{cleveref}
\usepackage{graphicx}
\usepackage{booktabs}
\usepackage[printonlyused]{acronym}

\usepackage{amsthm}

\crefname{ass}{assumption}{assumptions}
\crefname{theorem}{theorem}{theorems}
\crefname{prop}{proposition}{propositions}

\IEEEoverridecommandlockouts
\begin{document}
\input{commands}

\title{Collaborative Loco-Manipulation for Pick-and-Place Tasks with Dynamic Reward Curriculum}

\author{Tianxu An$^{\ast\dagger\ddagger}$, Flavio De Vincenti$^{\ast\dagger}$, Yuntao Ma$^\ddagger$, Marco Hutter$^\ddagger$, Stelian Coros$^\dagger$
\thanks{$^\ast$\textit{Tianxu An and Flavio De Vincenti contributed equally to this work.}}
\thanks{This work has received funding from the European Research Council (ERC) under the European Union’s Horizon 2020 research and innovation program (grant agreement No. 866480), and the Swiss National Science Foundation through SNF project grant 200021\_200644.}
\thanks{$^\dagger$The authors are with the Computational Robotics Lab (CRL), ETH Zurich, Switzerland. {\tt\small \{tianan, dflavio, scoros\}@ethz.ch}}%
\thanks{$^\ddagger$The authors are with the Robotic Systems Lab (RSL), ETH Zurich, Switzerland. {\tt\small \{mayun, mahutter\}@ethz.ch}}
}

\maketitle
\begin{abstract}
We present a hierarchical \ac{RL} pipeline for training one-armed legged robots to perform pick-and-place (P\&P) tasks end-to-end---from approaching the payload to releasing it at a target area---in both single-robot and cooperative dual-robot settings. We introduce a novel dynamic reward curriculum that enables a single policy to efficiently learn long-horizon P\&P operations by progressively guiding the agents through payload-centered sub-objectives. Compared to state-of-the-art approaches for long-horizon RL tasks, our method improves training efficiency by \SI{55}{\percent} and reduces execution time by \SI{18.6}{\percent} in simulation experiments. In the dual-robot case, we show that our policy enables each robot to attend to different components of its observation space at distinct task stages, promoting effective coordination via autonomous attention shifts. We validate our method through real-world experiments using ANYmal~D platforms in both single- and dual-robot scenarios. To our knowledge, this is the first \ac{RL} pipeline that tackles the full scope of collaborative P\&P with two legged manipulators.
\end{abstract}

\IEEEpeerreviewmaketitle

\input{sections/introduction}
\input{sections/related_work}
\input{sections/method}
\input{sections/results}
\input{sections/conclusion}




\bibliographystyle{plainnat}

\input{references.bbl}
\end{document}

%% file: commands.tex
\newcommand{\FDV}[1]{{\bf\textcolor{ForestGreen}{[FDV: #1]}}}
\newcommand{\TA}[1]{{\bf\textcolor{red}{[TA: #1]}}}

\newcommand{\X}{\mathbf{X}}
\newcommand{\x}{\mathbf{x}}
\newcommand{\e}{\mathbf{e}}
\newcommand{\p}{\mathbf{p}}
\newcommand{\q}{\mathbf{q}}

\newacro{RL}{reinforcement learning}
\newacro{PnP}[P\string&P]{pick-and-place}
\newacro{MA}{multi-agent}
\newacro{MARL}{multi-agent reinforcement learning}
\newacro{CLM}{collaborative loco-manipulation}
\newacro{MRS}{multi-robot system}
\newacro{EE}{end effector}
\newacro{DOF}{degree of freedom}
\acrodefplural{DOF}{degrees of freedom}
\newacro{LL}{low-level}
\newacro{HL}{high-level}
\newacro{FSM}{finite state machine}
\newacro{KL}{Kullback-Leibler}
\newacro{MPC}{model predictive control}
\newacro{MLP}{multilayer perceptron}
\newacro{CL}{curriculum learning}

%% file: sections/introduction.tex
\section{Introduction}
\label{sec:introduction}

\IEEEPARstart{M}{any} industries rely on human workers to perform physically demanding tasks such as lifting and transporting heavy loads through cluttered environments. Logistics and construction are among the most prominent examples, where workers engage in monotonous, repetitive \ac{PnP} operations with significant risks of physical injury. While advancements in hardware technology have made it possible for robots to traverse complex environments \cite{doi:10.1126/scirobotics.abc5986, miki2022learning} and dynamically manipulate rigid objects \cite{alma_bellicoso}, the necessary levels of autonomy and coordination for long-horizon mobile manipulation have yet to be effectively demonstrated in real-world scenarios.

\begin{figure}
    \centering
    \includegraphics[width=\linewidth]{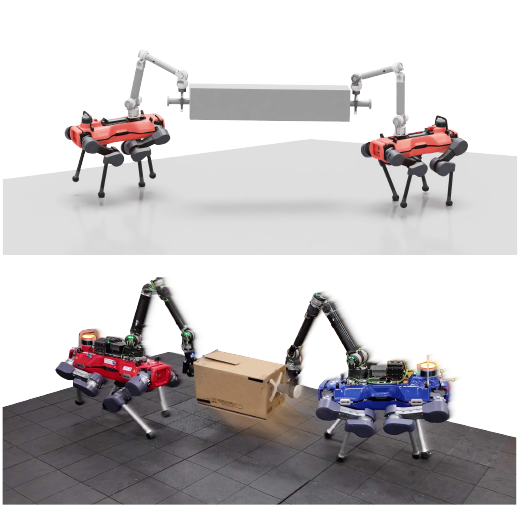}
    \caption{
    Collaborative loco-manipulation with ANYmal~D platforms in simulation (\textbf{top}) and on hardware (\textbf{bottom}). Our dual-robot reinforcement learning framework uses a dynamic reward curriculum to tackle long-horizon tasks composed of sequential sub-skills. Video recordings of the experiments are available in the supplementary materials.}
    \vspace{-0.5cm}
    \label{fig:teaser}
\end{figure}

We seek to develop a computationally efficient control architecture for the reliable end-to-end execution of \ac{PnP} tasks, from approaching the payload to releasing it at a target area. Since practical applications benefit greatly from highly adaptable, all-terrain mobile systems, we focus our study on legged machines equipped with a single arm, as depicted in \cref{fig:teaser}. When confronted with larger payloads, these robots must engage in collaborative manipulation to succeed. By leveraging cooperative behaviors, they can unlock new levels of functionality, achieving outcomes far beyond the reach of individual units, such as collaborative loco-manipulation with bulky, heavy objects \cite{flavio_mpc}.

Coordinating such complex systems poses significant challenges. Firstly, \ac{PnP} tasks often involve exploratory bottlenecks~\citep{needle_pnp}. For instance, robots must grasp an object before lifting it, making the grasping action a critical prerequisite for task progression. Secondly, in dual-robot systems, the environment is updated collectively, making it difficult for learning algorithms to account for the effects of individual robot actions. Lastly, the long-horizon nature of \ac{PnP} tasks, coupled with the system's inherent high dimensionality, calls for different control resolutions---ranging from fine-grained manipulation to high-level decision making---which significantly complicates the control problem. An effective control scheme must therefore ensure seamless, whole-body coordination while meeting the overall long-horizon objectives.

We present a hierarchical \ac{RL} framework that effectively tackles the above challenges by decomposing the \ac{PnP} task into a series of stages. We propose a dynamic reward curriculum that enables the robots to progressively master each sub-task. Our method allows for learning a decentralized \ac{HL} policy capable of accommodating either single- or dual-robot \ac{PnP} settings. Our \ac{HL} policy outputs base velocity and \ac{EE} pose commands as inputs to a separately trained \ac{LL} policy, while also controlling the gripper's opening and closing functions. Finally, we employ a teacher-student approach \cite{doi:10.1126/scirobotics.abc5986} to adapt the \ac{HL} policy for real-world deployment. Our contribution is threefold:

\begin{enumerate}
    \item We develop a hierarchical \ac{RL} controller that coordinates quadruped robots equipped with an arm and a gripper to perform \ac{PnP} tasks. Our learned policy simultaneously controls the robot's legs, arm, and gripper, enabling individual robots or collaborative pairs to transport bulky objects to designated locations.
    \item We introduce a dynamic reward curriculum that enables effective, sample-efficient learning of complex sequential behaviors in long-horizon, multi-stage problems.
    \item We conduct extensive simulation and real-world experiments, demonstrating that our policy outperforms prior state-of-the-art methods in both sample efficiency and robustness. To our knowledge, this is the first demonstration of collaborative loco-manipulation for autonomous \ac{PnP} using fully actuated single-armed quadrupeds.
\end{enumerate}

%% file: sections/related_work.tex
\section{Related Work}
\label{sec:related_work}

\subsection{Whole-Body Control of Single-Armed Quadruped Robots} The control of four-legged mobile manipulators requires the seamless coordination of locomotion and manipulation objectives. These tasks are distinct, yet interdependent, as precise \ac{EE} tracking often requires dynamic adjustments in posture, while stable locomotion must account for the shifting wrenches introduced by arm movements. While offering effective solutions, optimization-based \cite{alma_bellicoso, go_fetch, roloma, multicontact_planning} and differential dynamic programming approaches \cite{unified_mpc, unified_mpc_collision_free} are highly sensitive to model inaccuracies and estimation errors, limiting their applicability to structured, controlled environments. \citet{bayes_mpc} and \citet{rl_mpc} partially mitigate these issues by incorporating learning to either model system dynamics errors or predict arm-induced wrenches for smoother locomotion control. In contrast, fully integrated, whole-body \ac{RL} approaches show promise in overcoming both modeling and control limitations by jointly optimizing locomotion and manipulation~\cite{pan2024roboduetwholebodyleggedlocomanipulation, deep_whole_body, whole_body_force, visual_whole_body}.

Building on prior work by \citet{rl_mpc}, we implement our \ac{LL} controller with separate policies governing the leg and arm joints. Our method leverages fully learned policies to control all actuators within a unified framework.

\subsection{Collaborative Loco-Manipulation with Quadruped Robots} The transport of a payload with collaborative quadruped robots has long been investigated. Prior works involve groups of robots collaboratively pushing a large object with their bodies~\cite{2_quadruped_manip_via_loco, hierarchy_collab_manip}, carrying it on their backs~\cite{4_quadruped_payload}, or towing it using tethers~\cite{central_cable-towed_load} to reach a target location. These approaches often allow greater freedom of motion and exhibit weak interplay between robots. Another related line of work considers teams of quadruped robots carrying a payload rigidly attached to their bases, thus introducing additional holonomic constraints to the system \cite{3_quadruped_locomotion, 2_quadruped_locomotion, layer_coop_loco}. However, due to the rigid connection, these studies primarily focus on coordinated locomotion rather than direct manipulation of the object. Other works address this limitation by letting each robot hold the payload using a passive~\cite{mpc_passive_arm} or actuated robotic arm~\cite{flavio_mpc, mpc_coop_arm_carry}. Nonetheless, these approaches primarily address scenarios where the robots are already connected to the payload. To the best of our knowledge, our work is the first to tackle the entire scope of collaborative \ac{PnP}, from robots identifying and assigning handles on the payload, to grasping it, transporting it to a goal location, and ultimately releasing it in the desired spot.

\subsection{Multi-Stage Reinforcement Learning}

For complex, long-horizon tasks such as \ac{PnP}, most existing \ac{RL} approaches divide the complete problem into multiple stages and train a different controller for each~\citep{non_prehensile_manipulation, task_decomp_pnp}. The main limitation of these methods is that each policy is only aware of its corresponding sub-task, causing the overall performance to decline during stage transitions. To address these issues, some works introduce mechanisms to improve cross-stage consistency, such as optimizing a convex sum of critics from consecutive stages~\citep{multi_stage_coop_policy}, or transferring replay buffers to warm-start the current policy from a converged policy of the previous stage~\citep{needle_pnp}. Although effective, these methods still lack a holistic understanding of the full task sequence, negatively impacting the policy's performance.

Instead of employing multiple stage-specific policies, \citet{visual_whole_body} train a single policy to enable a single-armed quadruped robot to pick up objects from the ground. The authors design a stage allocator to identify the agent's current sub-task during training and assign corresponding stage-dependent rewards, ensuring a structured learning progression. Our approach adopts a similar stage allocator to define the training phases for the agents. Building on this, we introduce a dynamic reward curriculum that promotes smoother critic transitions during training, simplifies reward shaping, and significantly improves sample efficiency, while avoiding catastrophic forgetting of earlier stages.

\subsection{Curriculum Learning} Conventional \ac{CL} approaches typically address either a single skill with progressively increasing difficulty~\cite{hwangbo2019, luo2020, Rudin2021LearningTW, doi:10.1126/scirobotics.abc5986}, or train multiple sub-skills in an unguided fashion---either emergently through self-play~\cite{baker2020emergenttoolusemultiagent} or using teacher-student paradigms that do not explicitly account for the skill order~\cite{matiisen2020}. In contrast, we design our dynamic reward curriculum for tasks involving ordered sequences of sub-skills over extended time horizons.

%% file: sections/method.tex
\section{Method}
\label{sec:method}

We frame the \ac{PnP} problem by considering box-shaped objects with two custom-designed handles, as shown in \cref{fig:teaser}. Our goal is to enable either a single robot or a team of two to autonomously transport the object to a desired location by manipulating the handles alone.


\subsection{Hierarchical Policy}\label{subsec:hierarchical-policy}

\begin{figure}
    \centering
    \includegraphics[width=\linewidth]{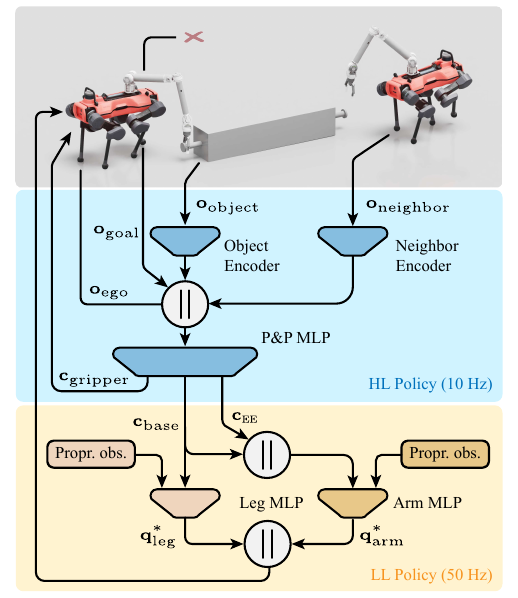}
    \caption{
        Overview of our hierarchical control architecture for collaborative \ac{PnP} tasks.
        The \ac{HL} policy observations consist of a concatenation of proprioceptive states ($\mathbf{o}_\mathrm{ego}$), the goal location ($\mathbf{o}_\mathrm{goal}$), an encoding of the object states ($\mathbf{o}_\mathrm{object}$), and an encoding of neighbor's states ($\mathbf{o}_\mathrm{neighbor}$). It outputs base velocity ($\mathbf{c}_\mathrm{base}$) and \ac{EE} pose ($\mathbf{c}_\textsc{ee}$) commands, which are processed by separate \acp{MLP} in the \ac{LL} policy along with proprioceptive data to compute joint angle targets for the legs ($\mathbf{q}_\mathrm{leg}^\ast$) and arm ($\mathbf{q}_\mathrm{arm}^\ast$). The \ac{HL} policy also outputs a binary command ($c_\mathrm{gripper}$) for the gripper. For single-agent settings, we set the encoding of neighbor states to zero.
    }
    \vspace{-0.5cm}
    \label{fig:overview}
\end{figure}

We adopt a bi-level control system consisting of a \ac{HL} policy producing base velocity, \ac{EE} pose, and gripper commands, and a \ac{LL} whole-body policy executing them at joint level. We train both policies using \ac{RL} by maximizing the discounted expected return $\mathop{\mathbb{E}}_\pi \left[\mathop{\sum}_{t=0}^{\infty} \gamma^t \mathcal{R}_t \right]$, where $\gamma \in [0, 1)$ is the discount factor, and $\mathcal{R}_t$ is the reward at time step $t$. We provide a schematic overview of our architecture in \cref{fig:overview}.

\subsubsection{Low-Level Policy}\label{subsec:ll-policy}

Aligned with previous work \cite{rl_mpc, Zhang2024LearningTO}, our \ac{LL} controller uses distinct policies to manage the leg and arm joints, $\q_\mathrm{leg}$ and $\q_\mathrm{arm}$, respectively. For locomotion, we adopt the learned policy by \citet{rl_mpc}.
On top of it, we train a separate manipulation policy that generates joint position targets for the 6 \acp{DOF} of the arm to track desired \ac{EE} pose and base velocity commands. 

\subsubsection{High-Level Policy}\label{subsec:hl-policy}

The \ac{HL} policy manages the task of approaching, picking up, and transporting a box to a designated goal position.
Its inputs include proprioceptive observations, object states, the goal position, and states of the potential collaborating robot. The policy outputs an $8$-dimensional vector of actions that serve as commands for the pre-trained \ac{LL} policy. It comprises the robot base velocity and \ac{EE} pose commands, with the latter defined by the position and yaw of the downward-oriented \ac{EE}. Finally, the \ac{HL} policy directly controls the opening and closing of the gripper.


\subsection{Dynamic Reward Curriculum}

Many long-horizon tasks like \ac{PnP} require agents to execute a fixed sequence of sub-tasks. Without this structure, agents tend to exploit undesirable shortcuts, such as kicking the object to the destination. To effectively capture the multi-stage nature of \ac{PnP} within a single \ac{HL} policy, we implement a \emph{dynamic reward curriculum}, which consists of a stage transition graph and an adaptive reward scaling rule.

\subsubsection{Stage Transition Graph} \label{subsubsec:stage-trans-graph}

\begin{figure*}[t]
    \centering
    \includegraphics[width=\textwidth]{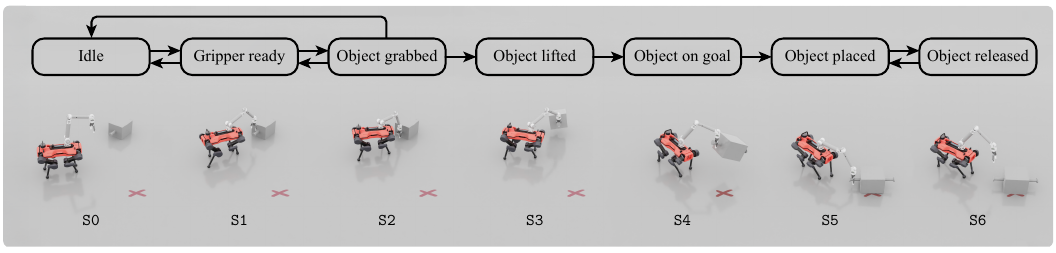}
    \caption{
    Stage transition graph, with arrows indicating possible transitions.
    }
    \label{fig:fsm}
\end{figure*}

We identify $7$ different stages and define the stage transition graph shown in \cref{fig:fsm}. At the start of each episode, the agents begin at stage 0 (\texttt{S0}) and progress sequentially through stages by meeting predefined transition conditions until task completion (\texttt{S6}). We strictly enforce forward transitions to ensure the agents master all intermediate skills. To facilitate the learning of recovery behaviors, we selectively enable backward transitions during the object's pickup and release stages, aiding the agents in refining their object interactions.



\subsubsection{Adaptive Reward Scaling}

We associate each stage $i$ with a set $J_i$ of reward functions $r_j$. Each stage-wise reward is designed to incentivize progression toward following stages, so that its maximization propels the agent to the next stage. Let $J \coloneqq \bigcup_i J_i$ denote the set of all stage-wise rewards. Then, we define the complete \ac{HL} reward function at episode $e$ as:
\begin{equation}\label{eq:hl-reward-function}
    \mathcal{R}_\textsc{hl}^e(\mathbf{s}) \coloneqq \mathcal{R}_\text{fixed}(\mathbf{s}) + \sum_{j \,:\, \exists r_j \in J} \rho_j^e \, r_j(\mathbf{s}) \,,
\end{equation}
where $\mathbf{s}$ is the full state of the system and $\mathcal{R}_\text{fixed}$ promotes overall desirable objectives---e.g., action regularization, collision avoidance, etc. The coefficients $\rho_j^e$ adapt dynamically across episodes to emphasize rewards relevant to the agent’s current stage while de-emphasizing others:
\begin{equation}\label{eq:dyn_rew_scale}
    \rho_j^e \coloneqq
\begin{cases}
\min\!\bigl(K \rho_j^{e-1}, \, \mu_j\bigr) & \text{if } r_j \in J_{\lambda(e-1)}\,, \\[6pt]
\max\!\bigl(\tfrac{1}{K}\rho_j^{e-1}, \, \mu_\textrm{min}\bigr) & \text{if } r_j \notin J_{\lambda(e-1)}\,,
\end{cases}
\end{equation}
where $K$ is a reward scale multiplier dictating the reactivity of the dynamic reward curriculum, the quantities $\mu_j > 0$ and $\mu_\textrm{min} > 0$ bound the scale of the $j$-th reward weight for episode $e$, and $\lambda(e-1)$ denotes the final stage reached by the robot at episode $e-1$. We remark that the weights $\rho_j^e$ remain constant during the same episode. Across episodes, they dynamically adjust as the policy internalizes the desired behaviors. As prescribed in the first line of \eqref{eq:dyn_rew_scale}, if the previous episode ends at stage $i$, i.e., $\lambda(e-1) = i$, then all the rewards corresponding to stage $i$ will get reinforced to help agents reach stage $i+1$; otherwise, their impact gets reduced.
It is worth noting that separate policy training schemes~\cite{needle_pnp, multi_stage_coop_policy} require at least as many rewards as our dynamic curriculum, resulting in a comparable number of hyperparameters. Our method adds only two, $K$ and $\mu_\textrm{min}$. All stage-wise rewards are available in \cref{tab:teacher_stage_rewards}.

Our approach improves sample efficiency by allowing the \ac{HL} policy to begin learning later stages before fully mastering earlier ones. At the same time, being exposed to reward signals from later stages fosters learning efficient forward-looking behaviors, such as nearing the target \emph{while} lifting the object. Finally, our unified training allows the policy to overcome unforeseen situations by reusing knowledge from different stages, such as re-grasping the handles if the object is dropped during transport, provided these events occur during training.

After a robot completes stage 6 for five consecutive training episodes, it enters a final stage with fixed reward scales across all stages and episodes. The final stage is designed to stabilize the training process and balance the optimization of both performance rewards $\mathcal{R}_\text{fixed}$ and stage rewards.



\begin{table}[t!]
  \centering
  \scriptsize
  \caption{Stage-Wise Rewards}
  \label{tab:teacher_stage_rewards}
  \renewcommand{\arraystretch}{0.85} 
  \begin{tabularx}{\linewidth}{>{\raggedright\arraybackslash}p{0.4cm}Xp{0.3cm}}
    \toprule
    \textbf{Stage} & \textbf{Reward} & $\mu_j$ \\
    \hline

    \texttt{S0} 
      & \textbullet\ \textit{gripper ready pose}: $\exp(-3\times(\mathrm{gripper\,xy\ dist.\ to\ handle})) \times |\cos(\text{gripper handle angle})|$ 
      & 2 \\[2pt]

    \texttt{S1} 
      & \textbullet\ \textit{gripper grasp pose}: $\exp(-10\times(\mathrm{gripper\,z\ dist.\ to\ handle})) \times |\cos(\text{gripper handle angle})|$ 
      & 20 \\
      & \textbullet\ \textit{object grabbed}: +1.0 if handle grasped 
      & 10 \\
      & \textbullet\ \textit{grab action}: (grasp probability) if grasped, (1 – probability) otherwise 
      & 0.2 \\[2pt]

    \texttt{S2} 
      & \textbullet\ \textit{object height}: $\min(\text{object height}, \text{max height})$ 
      & 100 \\[2pt]

    \texttt{S3} 
      & \textbullet\ \textit{object to goal vel}: $-\exp(-2\times\text{(object vel toward goal)})+1$ 
      & 50 \\
      & \textbullet\ \textit{object to goal dist}: $\exp\bigl(-2\times\frac{\text{object dist to goal}}{\text{initial dist to goal}}\bigr)$ 
      & 50 \\[2pt]

    \texttt{S4} 
      & \textbullet\ \textit{object drop}: $\max(0, \text{max obj height}-\text{obj height})$ 
      & 100 \\[2pt]

    \texttt{S5} 
      & \textbullet\ \textit{grab action in range}: $\max(0, 0.5-\text{grasp prob}) \times (\text{stage}\ge5)$ 
      & 100 \\[2pt]

    \texttt{S6} 
      & \textbullet\ \textit{object released in goal}: +1.0 if object released in goal 
      & 100 \\
      & \textbullet\ \textit{stage progress}: robot’s current stage 
      & 2 \\
      & \textbullet\ \textit{arm in default pose}: $\exp(-2\times\text{dist}(\text{arm dof}, \text{default}))$ 
      & 20 \\
    \bottomrule
  \end{tabularx}
\vspace{-0.5cm}
\end{table}

\subsection{Teacher-Student Distillation}

To bridge the sim-to-real gap, we implement a teacher-student training pipeline similar to \cite{doi:10.1126/scirobotics.abo0235}. We train the teacher policy using the dynamic reward curriculum, leveraging accurate measurements obtained from the simulation. Then, we train the student policy using fixed rewards and a policy distillation approach, which minimizes the \ac{KL} divergence between its action distribution and that of the teacher policy \cite{doi:10.1126/scirobotics.abo0235}. Unlike the teacher, the student policy operates with noisy measurements, lacks access to stage-related information, and must deduce the \ac{EE} pose from a history of the arm's \acp{DOF}.

The teacher policy may occasionally fail to recover from situations not encountered during training. For instance, due to the constraints imposed by the stage graph in \cref{fig:fsm}, the robots cannot transition back to \texttt{S0} after an external disturbance at \texttt{S4} causes the object to fall far from the target. In contrast, the student policy determines its stage solely based on external observations, enabling robots to recognize the new configuration as \texttt{S0} and autonomously restart the \ac{PnP} task.


%% file: sections/results.tex
\section{Results}
\label{sec:results}



\begin{figure*}
    \centering
    \includegraphics[width=0.9\textwidth]{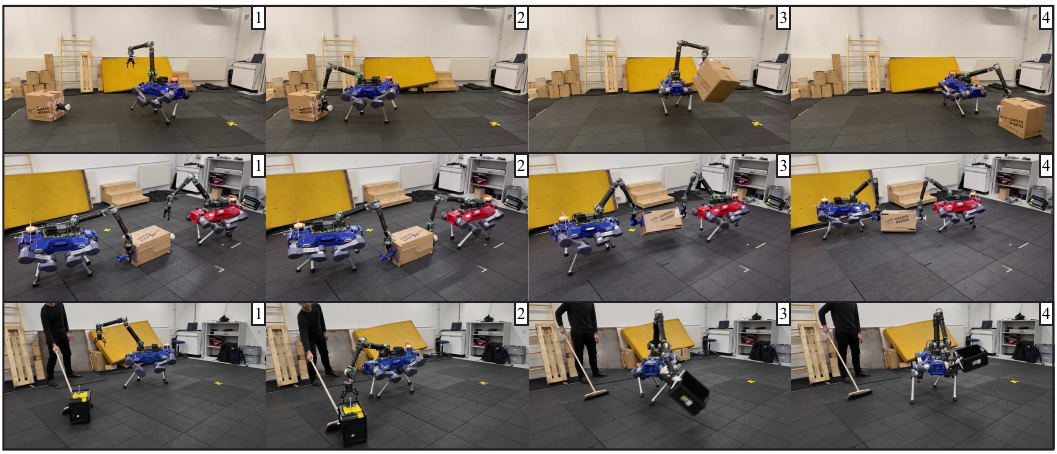}
    \caption{
    ANYmal~D robots performing \ac{PnP} tasks in single-robot (\textbf{top}) and dual-robot (\textbf{middle}) setups. The robot dynamically adapts to perturbations by re-grasping a displaced box (\textbf{bottom}).
    }
    \vspace{-0.2cm}
    \label{fig:hardware-test}
\end{figure*}

\subsection{Implementation Details.} We define the \ac{PnP} operation as successful if the box is delivered within \SI{0.5}{\meter} of the goal location. We trigger failure if any robot falls or if the task is not completed within \SI{45}{\second}. We randomly initialize the robots and objects around the goal position at distances ranging from $0.6$ to $5$ meters.
We also use boxes with different weights and shapes during training. For training single-robot \ac{PnP}, we use three different box assets with masses ranging from \SI{1.0}{\kilogram} to \SI{1.3}{\kilogram} and edge dimensions from \SI{0.22}{\meter} to \SI{0.45}{\meter}. For dual-robot \ac{PnP}, we use two different box assets with \SI{1.9}{\kilogram} and \SI{2.2}{\kilogram} weights, respectively, and edge dimensions from \SI{0.3}{\meter} to \SI{1.5}{\meter}. All box dimensions are randomly scaled with a factor between 0.8 and 1.2. For training, we use the \ac{MA} adaptation of the PPO algorithm \cite{ppo_algo} as described in \cite{10811859}. We conduct all our simulation experiments in Isaac Gym simulation environments \cite{Rudin2021LearningTW} with different starting configurations and object sizes.

\subsection{Hardware Experiments}\label{sec:hardware-experiments}

We test our policies in single- and dual-robot experiments using ANYmal~D platforms equipped with a 6-\ac{DOF} robotic arm DynaArm and the Robotiq 2F-140 gripper \cite{unified_mpc}. We task the robots with picking up and carrying either a \SI{1.5}{\kilogram} box (single-robot) or a \SI{2.0}{\kilogram} box (dual-robot) to a given target location. We provide snapshots of our experiments in \cref{fig:hardware-test}, where the goal position is marked on the ground using yellow tape. We implement multi-robot communications using the gRPC framework \cite{gRPC}, which serializes messages between a server and its clients distributed across different machines; we refer the reader to \cite{10811859} for implementation details. Finally, we estimate the pose of the box with respect to the robot bases through a motion capture system.

The single-agent policy succeeded in 8 of 9 trials, and the dual-agent policy in all 3 initial runs. In 3 additional trials, the robots delivered the box but stalled before drop-off due to mocap dropouts. We test the re-grasping behavior in the single-agent setup by moving the box around during the \ac{PnP} execution. As shown in the bottom row of \cref{fig:hardware-test}, the robot adapts dynamically by pursuing the displaced object, re-establishing its grasp, and completing the operation. These results highlight the system's reliability and robustness in realistic settings. We provide videos of the experiments in the supplementary materials.

\begin{figure}
    \centering
    \includegraphics[width=\linewidth]{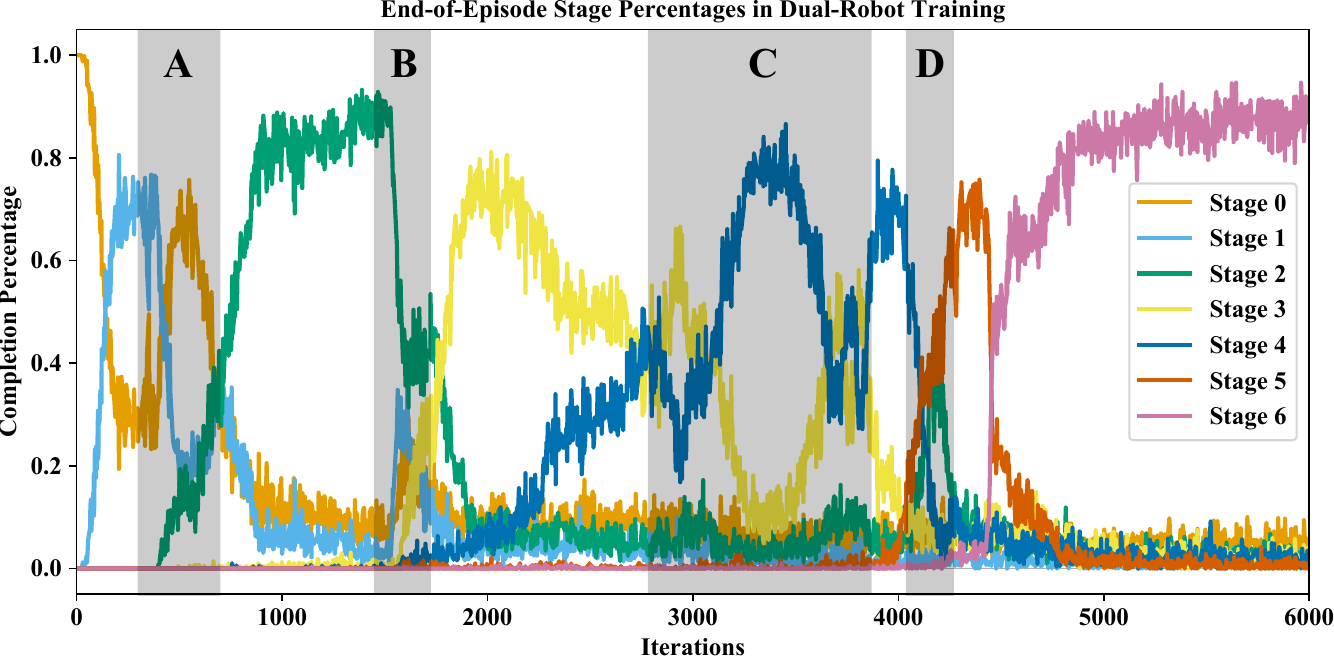}
    \caption{Stage progression in a dual-robot \ac{PnP} training with 2048 environments, showing the percentage ending in each stage across episodes. Four shaded sections highlight training falling back to previous stages for enhanced learning and cross-stage awareness. (\textbf{A}) The percentage of \texttt{S0} increases when reaching \texttt{S2}. (\textbf{B}) The percentages of \texttt{S0}, \texttt{S1}, and \texttt{S2} increase when reaching \texttt{S3}. (\textbf{C}) The percentage of \texttt{S3} increases between iteration 2800 and 3800 when reaching \texttt{S4}. (\textbf{D}) The percentage of \texttt{S2} increases when reaching \texttt{S5}.}
    \vspace{-0.5cm}
    \label{fig:stage_progression}
\end{figure}

\begin{figure*}[t]
    \centering
    \includegraphics[width=0.9\textwidth]{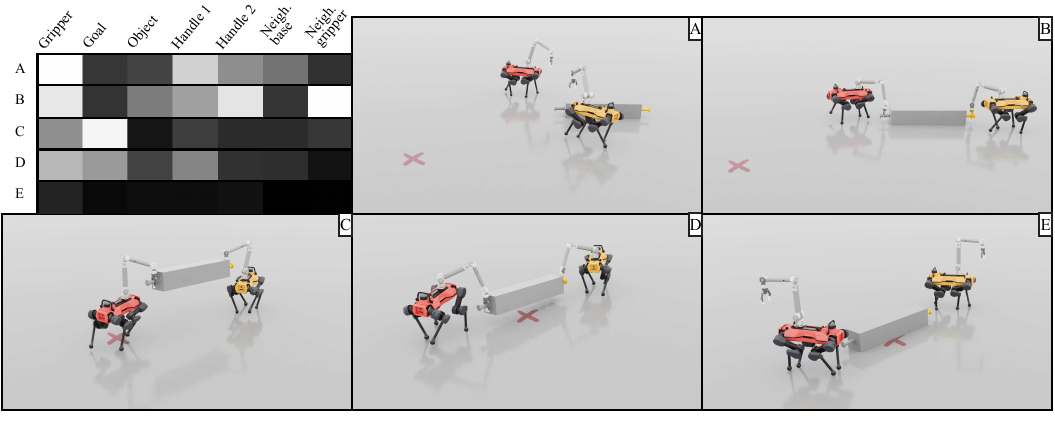}
    \caption{
    Saliency maps for selected observations at various stages of dual-robot \ac{PnP}. The ego robot is represented in red, with its neighbor in gold. Handle 1 is shown in gold, and handle 2 in grey. Lighter shades in the saliency map indicate more influential observations in guiding the policy's actions, while darker shades denote less relevant observations. The snapshots from A to E depict key stages of the task: reaching, lifting, transporting to the target location (at the red cross mark), laying, and releasing the object.
    }
    \vspace{-0.5cm}
    \label{fig:sensitivity-analysis}
\end{figure*}

\subsection{Analysis of Stage Progression During Training} \label{sec:stage-progression}

Training a single policy for long-horizon, multi-stage problems is challenging: the policy's action distribution must remain flexible enough for learning new tasks at later stages, yet not overfit to them so as to forget earlier ones. Moreover, the policy should be able to refine early-stage motions to facilitate later tasks, as well as acquire recovery strategies when those tasks fail. By analyzing the stage progression during a representative training in \cref{fig:stage_progression}, we show the benefits of our dynamic reward curriculum in addressing the above problems in dual-robot \ac{PnP}.

\subsubsection{Refining Motions for Later Stages}

In area A of \cref{fig:stage_progression}, stage progression reveals how motion refinement in earlier stages contributes to the successful completion of later ones. Specifically, with reference to \cref{fig:fsm}, both robots begin at \texttt{S1} (``Gripper ready'') and are learning to grasp the handles (\texttt{S2}) after having aligned their grippers above them. However, lowering the grippers without precise alignment may violate the condition of \texttt{S1}, forcing a backward transition to \texttt{S0} (``Idle''). When this occurs, the robots are required to re-optimize the \texttt{S0} rewards, thus re-establishing correct alignment above the handles. This repeated emphasis leads to smoother grasping motions, ensuring the grippers approach the handles from above rather than from the side, thereby avoiding collisions and deadlocks during execution.

\subsubsection{Learning Recovery Behaviors}

Our experiments revealed the emergence of recovery behaviors following disruptive events, as discussed in \cref{sec:hardware-experiments}. This is especially visible in area B of \cref{fig:stage_progression}: when attempting to lift the object from \texttt{S2} (``Object grabbed''), unstable grasp configurations---still considered valid for \texttt{S2}---often cause the object to slip, resulting in a fallback to \texttt{S0}. However, the increasing percentages in \texttt{S1}--\texttt{S2} indicate that the robots learn to recover by re-attempting the grasp after such failures. Furthermore, the comparable proportions of episodes across stages \texttt{S0}--\texttt{S3} (\SI{20}{\percent}--\SI{40}{\percent}) suggest that the combined rewards for approaching, grasping, and lifting promote the rejection of unstable grasping strategies, producing smooth, forward-looking motions.

\subsubsection{Overcoming Stage Bottlenecks}

Having reached \texttt{S4} (``Object on goal''), learning \texttt{S5} (``Object placed'') is particularly challenging, as it requires two robots to coordinate the lowering motion, directly conflicting with the previously learned lifting behavior. PPO struggles in this setting because the necessary actions lie in a low-probability region of the learned distribution. As shown in area C of \cref{fig:stage_progression}, the consequent lack of immediate rewards for the \texttt{S4}$\rightarrow$\texttt{S5} transition makes reaching \texttt{S4} appear unprofitable. Thus, the optimizer may maximize the overall return by avoiding \texttt{S4} altogether and repeatedly exploiting \texttt{S3}$\rightarrow$\texttt{S4} rewards instead. Our dynamic reward curriculum mitigates this regression by adaptively scaling up the \texttt{S3} rewards, thereby pushing the policy past the bottleneck and enabling steady progress, as evidenced by the shifts around 2800 and 3800 training iterations in \cref{fig:stage_progression}.

\subsubsection{Mitigating Forgetting}

Once \texttt{S5} is learned, policies in area D of \cref{fig:stage_progression} tend to forget previously acquired lifting motions, causing more episodes to terminate at \texttt{S2}. By re-emphasizing lifting rewards, our curriculum enables robots to consistently perform both lifting at \texttt{S2} and placing at \texttt{S4}---a difficult pair of opposing motions for a single \ac{RL} policy.

\subsection{Sensitivity Analysis}\label{sec:sensitivity-analysis}

We conduct a sensitivity analysis to understand how robots allocate their focus on neighboring agents, the object, and the goal position \cite{10811859}. Specifically, at each time step we compute the derivative of the $\ell_2$ norm of the action vector with respect to all input observations, including those of the ego robot, the goal, the object, and the neighboring robot. A small gradient magnitude means that perturbing an observation produces minimal changes in action, indicating a lower impact on the robot's decision-making process; conversely, a large magnitude suggests a greater impact.

The saliency maps in \cref{fig:sensitivity-analysis} show that the robots dynamically shift their attention to different elements of the environment depending on the task phase. During the approaching stage (A), the ego robot attends to both object handles and the neighboring robot's base to decide on handle assignment. At the grasping and lifting stages (B), the ego robot focuses the most on its gripper, the handle it is grasping, and the neighboring robot's gripper, ensuring coordinated lifting. As the robots transport the object to the goal (C), most of the attention shifts to the gripper and the goal position. When nearing the release stage (D), the ego robot pays little direct attention to its neighbor and instead focuses on the handle held by the partner (Handle 1). This suggests it learns to infer the neighbor’s motion through the handle, exploiting their physical coupling. Such reactivity enables the robots to respond effectively to changing circumstances, enhancing their coordination, efficiency, and overall task success.

\subsection{Ablation Study on Dynamic Reward Curriculum}\label{sec:ablation}

\begin{figure}
    \centering
    \includegraphics[width=\linewidth]{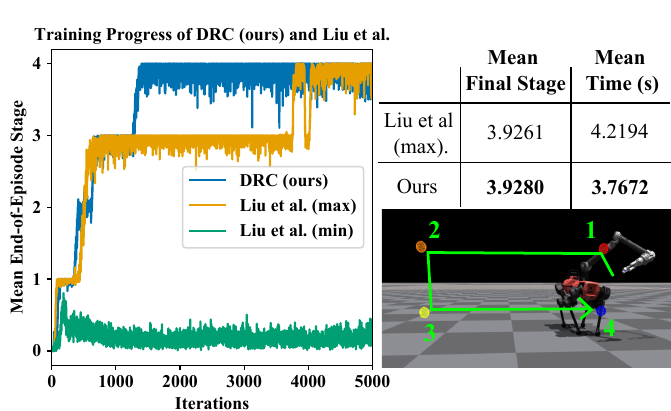}
    \caption{
    Ablation of the dynamic reward curriculum on a 4-stage \ac{EE} position-tracking task. The robot sequentially tracks points 1--4 with its gripper.
    }
    \vspace{-0.2cm}
    \label{fig:ablation_exp}
\end{figure}

We conduct an ablation study of the dynamic reward curriculum on a 4-stage \ac{EE} position tracking problem---see \cref{fig:ablation_exp}. Although each stage involves the same type of task (tracking positions), the challenge lies in the long-horizon nature of the problem. As robots progress to later stages, their rewards become harder to tune because earlier stage rewards grow too large and dominate gradient optimization. The ablated method ``Liu et al. (max)'' follows \citet{visual_whole_body}: once a stage is completed, its rewards are deactivated and set to the maximum values to encourage agents to complete more stages.
\cref{fig:ablation_exp} shows that ``Liu et al. (max)'' gets stuck at \texttt{S3} for about 3000 iterations, since large rewards from earlier stages make it hard for the optimizer to pick up the initially small rewards for \texttt{S4}.

As another variant, we set rewards from completed stages to the minimum value so the policy might more easily pick up new rewards; we call this ablated model ``Liu et al. (min).'' This version fails to pass even \texttt{S1}, as the policy prefers exploiting mastered rewards rather than transitioning to new stages with no immediate return. In summary, for long-horizon multi-stage \ac{RL}, both ablated models suffer from difficult reward tuning, local minima, and poor sample efficiency. In contrast, our dynamic reward curriculum adaptively adjusts stage rewards during training, mitigating these issues and yielding higher task success and faster completion times, as shown in \cref{fig:ablation_exp}.

\begin{figure}
    \centering
    \includegraphics[width=\linewidth]{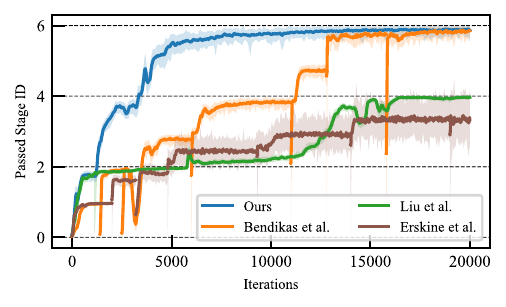}
    \caption{
        Mean end-of-episode stage number for single-agent \ac{PnP} across different approaches. The \textit{y}-axis indicates the average stage number reached by the \ac{HL} policy at the end of an episode, while the \textit{x}-axis shows the number of iterations required for the policy to reach that stage.
    }
    \vspace{-0.5cm}
    \label{fig:policy-stage_curve}
\end{figure}

For completeness, we also train the ablated model (\citet{visual_whole_body}) on the single-agent \ac{PnP} problem. As shown in \cref{fig:policy-stage_curve} and \cref{tab:com_exp_metric},it stalls after reaching \texttt{S4}, for the same reasons discussed above.


\subsection{Benchmark Comparison}\label{sec:benchmarks}

We benchmark our complete pipeline against two representative state-of-the-art methods that train separate policies for multi-stage \ac{RL} problems---\citet{needle_pnp} and \citet{multi_stage_coop_policy}---on the single-agent \ac{PnP} task. Both approaches incorporate critic functions of adjacent stages to encourage cross-stage coherence when training separate policies. The key distinction lies in the direction of this integration: \citeauthor{needle_pnp} include critics from earlier stages and train each policy to execute all preceding sub-tasks, using only the final-stage policy at inference time. In contrast, \citeauthor{multi_stage_coop_policy} optimize each policy solely for its assigned stage while incorporating the next-stage critic to guide training.

\begin{table}[t]
    \caption{Performance metrics of different methods averaged over 500 trials in simulation.}
    \label{tab:com_exp_metric}
    \centering
    \renewcommand{\arraystretch}{0.8} 
    \begin{tabular}{@{}lll@{}} 
      \toprule
       & \textbf{Success Rate} (\%) & \textbf{Completion Time} (s) \\
      \midrule
       \textit{Single-Robot \ac{PnP}} & & \\
       \quad Ours & \textbf{93.6} & \textbf{10.78 $\pm$ 0.34} \\
       \quad \citet{visual_whole_body} & 0 & - \\
       \quad \citet{needle_pnp} & 91.2 & 13.25 $\pm$ 0.41 \\
       \quad \citet{multi_stage_coop_policy} & 52.8 & 16.00 $\pm$ 0.87 \\
      \midrule
       \textit{Dual-Robot \ac{PnP}} & & \\
       \quad Ours & \textbf{90.3} & \textbf{10.84 $\pm$ 0.47} \\
      \bottomrule
    \end{tabular}
    \vspace{-0.5cm}
\end{table}

\subsubsection{Bendikas et al.'s approach}

As shown in \cref{tab:com_exp_metric}, \citeauthor{needle_pnp}'s approach achieves a competitive average success rate of \SI{91.2}{\percent}, which is close to ours. However, since each policy is trained without knowledge of later stages, behaviors learned at earlier stages often lead to uncoordinated actions that are difficult to adjust as training progresses. For example, we observe that robots trained with our dynamic reward curriculum exhibit smooth, stable transporting motions, whereas robots trained using \citeauthor{needle_pnp}'s method extend their arm sideways while holding the object, risking imbalance during transport---see the supplementary video for a visual comparison. This issue arises because the lifting policy, trained without accounting for the transport of the lifted object to the goal, carries over clumsy behaviors to later stages, hindering base balance. These suboptimal behaviors help explain the longer completion times compared to our method. Additionally, as shown in \cref{fig:policy-stage_curve}, \citeauthor{needle_pnp}'s approach is less sample-efficient than ours, as it requires each policy to converge before the next one can start training. In contrast, our pipeline allows the agents to refine later-stage actions even before earlier stages are fully learned.

\subsubsection{Erskine et al.'s approach}

The low success rate in \cref{tab:com_exp_metric} (\SI{52.8}{\percent}) is consistent with \citeauthor{multi_stage_coop_policy}'s findings, whereby agents struggle to learn later-stage policies if earlier-stage policies end up in challenging initial states. This is also the case in our study: at \texttt{S0}, the robot learns a policy to keep its body close to the object in a ready-to-grasp pose, which hampers the learning of the lifting policy at \texttt{S3}. Unlike our approach and that of \citeauthor{needle_pnp}, executing the full motion with separate policies offers no guarantee of convergence. This leads to inconsistent performance, as shown in \cref{fig:policy-stage_curve}, where the variance of the end-of-episode stages, indicated by the shaded area around the mean curve, increases noticeably---especially when training policies for later stages.

\subsection{Limitations}
\label{sec:limitations}

Collaborative multi-robot \ac{PnP} requires defining stage-wise behaviors to prevent undesired shortcuts. While intuitive, reward design can be time-consuming, involving numerous terms and tedious tuning. Future work could automate this process, e.g., via meta-learning or foundation models.

In simulation, most failures occur when robot grippers get stuck by the object handle, either due to poor grasping motions, such as sideways grasps, or the object being perturbed by another robot's grasp. Although our policy can deal with most cases with over \SI{90}{\percent} success rate, this highlights the difficulty of collaborative manipulation.

Our method also depends on accurate object pose and goal measurements. In real-world settings, sensor noise or mocap dropouts can cause failures, as seen in our videos. More robust policies fusing multiple noisy sensors could mitigate this issue.

Finally, hardware tests reveal limited compliance: under communication delays, robots resist each other instead of adapting. Future work may explore compliant control or learned impedance modulation to enhance adaptability and robustness in collaborative manipulation.


%% file: sections/conclusion.tex
\section{Conclusion}
\label{sec:conclusion}


In this work, we presented a hierarchical \ac{RL} framework for collaborative \ac{PnP} with single- and dual-robot systems. Our bi-level control architecture decouples \ac{HL} planning from \ac{LL} execution, allowing robots to approach, grasp, and transport voluminous objects. We train the \ac{HL} policy with a dynamic reward curriculum that guides progression through sub-tasks over long time horizons while boosting sample efficiency. Finally, we applied teacher-student distillation to bridge the sim-to-real gap, enabling effective transfer of policies to hardware.